%% file: main.tex
\documentclass[letterpaper, 10pt, conference]{ieeeconf}

\input{preamble}

\IEEEoverridecommandlockouts  
\usepackage[english]{babel}

\usepackage{graphicx} 
\usepackage{amsmath}
\usepackage{amssymb}
\usepackage{amsthm}
\usepackage{algorithm}
\usepackage{algpseudocode}
\usepackage{amsfonts}
\usepackage{float}
\usepackage{mathrsfs}
\usepackage{booktabs}
\usepackage{array}
\usepackage[nodisplayskipstretch]{setspace}
\makeatletter
\def\BState{\State\hskip-\ALG@thistlm}
\makeatother
\usepackage[capitalize,nameinlink,noabbrev]{cleveref} 

\theoremstyle{definition}

\newcolumntype{M}[1]{>{\centering\arraybackslash}m{#1}}
\usepackage{pifont}
\usepackage{multirow}
\usepackage{tabularray}
\usepackage{cite}
\usepackage{algpseudocode}

\newcommand{\RNum}[1]{\uppercase\expandafter{\romannumeral #1\relax}}
\usepackage{bigints}

\newcommand{\genTISD}{\textbf{\texttt{genTISD}}}
\newcommand{\Eclares}{\textbf{\texttt{Eclares}}}
\newcommand{\eware}{\textbf{\texttt{eware}}}
\algrenewcommand\textproc{}

\algrenewcommand\algorithmicrequire{\textbf{Input:}}
\algrenewcommand\algorithmicensure{\textbf{Output:}}

\hypersetup{
  colorlinks,
  citecolor=teal, 
  linkcolor=teal,
  urlcolor=purple}

\author{Kaleb Ben Naveed$^{1}$,  Devansh Agrawal$^{2}$, Christopher Vermillion$^{3}$ and Dimitra Panagou$^{1,2}$%
\thanks{$^{*}$The authors would like to acknowledge the support of the National Science Foundation (NSF) under grant no. 2223845 and grant no. 1942907.}
\thanks{$^{1}$Department of Robotics, University of Michigan, Ann Arbor, MI, 48109 USA. {\{\tt\small kbnaveed@umich.edu\}}}%
\thanks{$^{2}$ Department of Aerospace Engineering, University of Michigan, Ann Arbor, MI, 48109 USA. }
\thanks{$^{3}$ Department of Mechanical Engineering, University of Michigan, Ann Arbor, MI, 48109 USA. }
}  

\title{\LARGE \bf
\texttt{Eclares}: Energy-Aware Clarity-Driven Ergodic Search
} 

\newcommand\footnoteref[1]{\protected@xdef\@thefnmark{\ref{#1}}\@footnotemark}

\begin{document}

\maketitle
\thispagestyle{empty}
\pagestyle{empty}

\begin{abstract}
\input{subsections/abstract.tex}
\end{abstract}



\section{Introduction}

\input{subsections/Intro}


\section{Preliminaries}
\input{subsections/prelim}

\section{Problem Formulation}
\input{subsections/problem_formulation}

\section{Energy Aware Clarity Driven Ergodic Search (\textbf{\texttt{Eclares}})}
\input{subsections/Eclares}

\input{subsections/experiments}
\input{subsections/conclusion}

\nocite{*}
\bibliographystyle{IEEEtran}
\bibliography{biblio.bib}

\appendix
\input{subsections/Appendix}

\end{document}

%% file: preamble.tex
\usepackage{graphics}
\usepackage{epsfig}

\usepackage{cite}
\usepackage[dvipsnames, svgnames]{xcolor}
\usepackage{xurl}
\usepackage{hyperref}
\hypersetup{
    colorlinks,
    linkcolor={red!50!black},
    citecolor={blue!50!black},
    urlcolor={blue!80!black}
}
\usepackage{bm}
\usepackage{breqn}
\usepackage{amssymb}
\usepackage{tabularx}
\usepackage{booktabs}
\usepackage{dcolumn}
\newcolumntype{d}[1]{D..{#1}}
\newcolumntype{C}{>{\centering\arraybackslash}X}
\usepackage{soul}

\usepackage{amsthm}

\newcommand{\reals}{\mathbb{R}}
\newcommand{\R}{\reals}
\newcommand{\Rnonneg}{\reals_{\geq 0}}
\newcommand{\Rplus}{\reals_{>0}}

\newcommand{\naturals}{\mathbb{N}}
\renewcommand{\S}{\mathbb{S}}

\newcommand{\pd}{\S_{++}}

\newcommand{\Acal}{\mathcal{A}}
\newcommand{\Bcal}{\mathcal{B}}
\newcommand{\Ccal}{\mathcal{C}}

\newcommand{\Pcal}{\mathcal{P}}

\newcommand{\Ucal}{\mathcal{U}}
\newcommand{\Xcal}{\mathcal{X}}

\newcommand{\Zcal}{\mathcal{Z}}

\newcommand{\eqn}[1]{\begin{align} #1 \end{align}}

\newcommand{\norm}[1]{\left\Vert #1 \right \Vert}



%% file: subsections/abstract.tex
Planning informative trajectories while considering the spatial distribution of the information over the environment, as well as constraints such as the robot's limited battery capacity, makes the long-time horizon persistent coverage problem complex. Ergodic search methods consider the spatial distribution of environmental information while optimizing robot trajectories; however, current methods lack the ability to construct the target information spatial distribution for environments that vary stochastically across space and time. Moreover, current coverage methods dealing with battery capacity constraints either assume simple robot and battery models, or are computationally expensive. To address these problems, we propose a framework called \texttt{Eclares}, in which our contribution is two-fold. 1) First, we propose a method to construct the target information spatial distribution for ergodic trajectory optimization using clarity, an information measure bounded between $[0,1]$. The clarity dynamics allows us to capture information decay due to lack of measurements and to quantify the maximum attainable information in stochastic spatiotemporal environments. 2) Second, instead of directly tracking the ergodic trajectory, we introduce the energy-aware (\texttt{eware}) filter, which iteratively validates the ergodic trajectory to ensure that the robot has enough energy to return to the charging station when needed. The proposed \texttt{eware} filter is applicable to nonlinear robot models and is computationally lightweight. We demonstrate the working of the framework through a simulation case study.
\href{https://github.com/kalebbennaveed/Eclares.git}{[Code]}\footnote{Code: https://github.com/kalebbennaveed/Eclares.git}\href{https://youtu.be/1ZCgxlHitzk}{[Video]}\footnote{Video: https://youtu.be/1ZCgxlHitzk}
\vspace{-2mm}


%% file: subsections/Intro.tex
Autonomous robots are widely used in tasks that involve data acquisition over long time horizons, such as search-and-rescue \cite{search_rescue1, search_rescue2}, characterization of ocean currents such as the Gulf Stream to study cross-shelf exchanges \cite{gulf1, gulf2}, water body exploration \cite{water_exploration1, water_exploration2} and wildfire monitoring \cite{julian2019distributed}. Coverage planning for data acquisition poses diverse challenges, including considering the spatial distribution of environmental information while optimizing search trajectories \cite{Mezic_Ergodic, Dressel_Ergodic, dong2023time}. This is important because some regions have higher density of information as compared to others. Other challenges involve persistently monitoring a \textit{spatiotemporal environment} that varies across space and time \cite{Dressel_Ergodic, regret_benhayden, pratap_persistent, julian2019distributed}, and ensuring \textit{task persistence} by enabling the robot to return to the charging station and recharge when/as needed during persistent coverage tasks \cite{persis_cbf1, persis_cbf3, persis_cbf4, will2}. In this paper, we consider these problems and propose a solution framework \textbf{\texttt{Eclares}}. Moreover, we also consider the case when a spatiotemporal environment has process uncertainty, i.e., we consider a class of \textit{stochastic spatiotemporal environments}. Probabilistic methods can be used to compute estimates of the environmental variation. However, in general the stochastic variations in the environment can result in information decay without continuous monitoring.

Recent advances in ergodic search \cite{Mezic_Ergodic, Dressel_Ergodic, abraham2021ergodic, cmu_ergodic, dong2023time} consider the spatial distribution of the information over the environment while generating coverage trajectories. This is acheieved by constructing a  \emph{target information spatial distribution} (TISD) over the environment and then adjusting the time spent in specific regions according to the TISD. However, these methods either assume \textit{spatiostatic environment} \cite{Mezic_Ergodic, dong2023time}, which only varies through space, or spatiotemporal with known variation model \cite{Dressel_Ergodic, multiobjectiveergodic, Candela_thesis}. Therefore, these methods can not incorporate potential information decay, which will result from the environment's stochastic variation across space and time. To address this problem, we use clarity \cite{clarity}, an information measure bounded between $[0,1]$, to construct the TISD. The proposed method to construct the TISD captures information decay due to a lack of measurements, and quantifies the maximum attainable information in stochastic spatiotemporal environments. 

In order to ensure persistent monitoring of a stochastic spatiotemporal environment for a long time horizon, the robot's limited battery capacity must be taken into consideration so that it can return to a charging station when needed. Most of the recent work on task persistence mainly either uses control barrier function (CBF)-based methods \cite{ames2016control, persis_cbf1, persis_cbf3, persis_cbf4} or Hamilton-Jacobi reachability based methods \cite{bansal2017hamilton, will2, clarity}. CBF-based methods are computationally efficient; however, they only assume simple robot and battery models. Alternatively, reachability-based methods can handle nonlinear systems and are robust with respect to external disturbances, but they are mostly used offline, suffer from curse of dimensionality, and are computationally expensive. The proposed online filter \textbf{\texttt{eware}} inspired by \cite{gatekeeper} addresses these problems. \textbf{\texttt{eware}} iteratively validates the ergodic trajectory based on the robot's current battery levels, ensuring that the robot is aware of its remaining battery energy during the coverage task and returns to the charging station before the energy is depleted. In summary, the main contributions of the proposed framework $\Eclares$ include:
\begin{itemize}
    \item Clarity-driven method to generate TISD, Which incorporates potential information decay and quantifies maximum attainable information in a stochastic spatiotemporal environment;
    \item \textbf{\texttt{eware}} filter which iteratively validates the generated ergodic trajectory to ensure that the robot has enough energy to return to the charging station when needed.
\end{itemize}

The paper is organized as follows: \cref{sec:prelim} introduce preliminaries, \cref{sec:problem_formulation} formulates the problem statement, \cref{sec:ECLARES} present the proposed solution framework \textbf{\texttt{Eclares}} and \cref{sec:experiments} discuss simulation setup and results.

%% file: subsections/prelim.tex
In this section, we outline the notation used throughout the paper, provide an overview of the ergodic search methods, and introduce clarity as an information measure. 
\label{sec:prelim}
\subsubsection{\textbf{Notation}}Let $\naturals = \{ 0, 1, 2, ... \} $. Let $\mathbb{R}$, $\mathbb{R}_{\geq 0}$, $\mathbb{R}_{> 0}$ be the set of reals, non-negative reals, and positive reals respectively. Let $\mathcal{S}^{n}_{++}$ denote set of symmetric positive-definite in $\mathbb{R}^{n \times n}$. Let $\mathcal{N}(\mu, \Sigma)$ denote a normal distribution with mean $\mu$ and covariance $\Sigma \in \mathcal{S}^{n}_{++}$. The space of continuous functions $f: \Acal \to \Bcal$ is denoted $C(\Acal, \Bcal)$. The $Q \in \pd^{n}$ norm of a vector $x \in \R^n$ is denoted $\norm{x}_Q = \sqrt{x^T Q x}.$

\subsubsection{\textbf{Dynamics Model}}
Consider the continuous-time system dynamics comprising the robot and battery discharge dynamics:

\begin{equation}
\label{eqn:aug_ctrlaffine_sys}
     \Dot{\chi} =
    \begin{bmatrix}
        \Dot{x} \\
        \Dot{e} 
    \end{bmatrix}
    =
    f(\chi, u) \\
    =
    \begin{bmatrix}
        f_r(x, u) \\
        f_e(e, x, u) 
    \end{bmatrix}
\end{equation}
where  $\chi = \begin{bmatrix}{x}^T, & e\end{bmatrix}^T \in \Zcal \subset \R^{n+1}$ is the system state consisting of the robot's state $x \in \mathcal{X} \subset \mathbb{R}^{n}$ and robot's State-of-Charge (SoC) $e \in \mathbb{R}_{\geq 0}$. $u \in \mathcal{U} \subset \mathbb{R}^{m}$ is the control input, $f: \Zcal \times \Ucal \rightarrow \mathbb{R}^{n+1}$ defines the continuous-time system dynamics, $f_r: \Xcal \times \Ucal \rightarrow \mathbb{R}^n$ define robot dynamics and $f_e: \mathbb{R}_{\geq 0} \times \Xcal \times \Ucal \to  \mathbb{R}$ define battery discharge dynamics. 



\subsubsection{\textbf{Ergodic Search}}

\label{sec:prelim_ergodic}

Ergodic search~\cite{Mezic_Ergodic, Dressel_Ergodic} is a technique to generate trajectories $x: [t_0, T] \to \mathcal{X}$ that cover a rectangular domain  $\mathcal{P} = [0, L_1] \times \cdots [0, L_s] \subset \mathbb{R}^s$, matching a specified \emph{target information spatial distribution} (TISD) $\phi : \Pcal \to \R$, where $\phi(p)$ is the density at $p \in \mathcal{P}$. 
Moreover, the spatial distribution of the trajectory $x(t)$ is defined as
\begin{equation}
    c(x(t), p) = \frac{1}{T-t_0}\int_{t_0}^{T} \delta(p - \Psi(x(\tau))) d\tau
\end{equation}
where $\delta: \Pcal \to \R$ is the Dirac delta function and $\Psi: \Xcal \to \Pcal$ is a mapping such that $\Psi(x(\tau))$ is the position of the robot at time $\tau \in [t_0, T]$. In other words, given a trajectory $x(t)$, $c(x(t), p)$ represents the fraction of time the robot spends at a point $p \in \Pcal$ over the interval $[t_0, T]$. Then, the \emph{ergodicity} of $x(t)$ w.r.t to a TISD $\phi$ is
\eqn{
\Phi(x(t), \phi) = \norm{ c - \phi}_{H^{-(s+1)/2}}
}
where $\norm{\cdot}_{H^{-(s+1)/2}}$ is the Sobolev space norm defined in~\cite{Mezic_Ergodic}, i.e., $\Phi$ is a function space norm measuring the difference between the TISD $\phi$ and the spatial distribution of the trajectory $c$. In~\cite{Dressel_Ergodic} a Projection-based Trajectory Optimization (PTO) method is proposed to generate ergodic trajectories. It optimizes (over the space of trajectories $x(t) \in C([t_0, T], \Xcal), u(t) \in C([t_0, T], \Ucal)$)
\begin{equation}
    \label{eq:ergodic_optimzation_cont}
    \begin{aligned}
    \min_{x(t), u(t)} \quad & \Bigl[\Phi(x(t), \phi) + J_b(x(t)) + \int^{T}_{t_0} \norm{u(\tau)}^2 d\tau \Bigr]\\
    \textrm{s.t.} \quad & \Dot{x} = f(x, u) \\
    & x_{0} = x(t_0) 
    \end{aligned}
\end{equation}
where $J_b$ penalizes trajectories that leave the domain, and $x_{0}$ is the robot's state at time $t_0$. The algorithm uses Fourier decomposition of $c, \phi$ to numerically evaluate $\Phi$ and its gradients with respect to $x(t)$ and $u(t)$. This enables an efficient gradient descent algorithm to optimize \eqref{eq:ergodic_optimzation_cont}. In this paper, we do not modify the algorithm in~\cite{Dressel_Ergodic}. Instead, we focus on developing a strategy to systematically construct the TISD $\phi$ given the environment model and the robot's sensing model, as discussed later.

\subsubsection{\textbf{Clarity}}
\label{sec:prelim_clarity}

Consider a stochastic variable (quantity of interest) $h\in \R$ governed by the process and output (measurement) models:
\begin{subequations}
\eqn{
    \Dot{h} &= w(t),   &&w(t) \sim \mathcal{N} (0, Q) \label{eqn: quantity_of_interest_point}\\ 
    y &= C(x)h+ v(t),  &&v(t) \sim \mathcal{N} (0, R)   \label{eqn: quantity_of_interest_measurement_point}
}
\end{subequations}
where $Q \in \Rnonneg$ is the known variance associated with the process noise, $y \in \R$ is the measurement, $C: \Xcal \to \R$ is the mapping between robot state and sensor state\footnote{C is set to 1 when sensed by the robot and 0 otherwise.}, and $R \in \R$ is the known variance of the measurement noise. 

In \cite{clarity}, we introduced the notion of clarity $q$ of the random quantity $h$, which lies between $[0,1]$ and is defined such that $q = 0$ represents $h$ being unknown, and $q = 1$ corresponds to $h$ being completely known. The clarity dynamics for the subsystem \eqref{eqn: quantity_of_interest_point}, \eqref{eqn: quantity_of_interest_measurement_point} are given by \cite{clarity}
\begin{equation}
    \label{eqn:clarity_dynamics}
    \Dot{q} = \frac{C(x)^2}{R}(1 - q)^2 - Qq^2
\end{equation}
If $C(x) = C$ is constant, \eqref{eqn:clarity_dynamics} admits a closed-form solution for the initial condition $q(0) = q_0$:
\begin{equation}
    \label{eqn:clarity_dynamics_closed_form}
    q(t; q_0) = q_{\infty} \left(1 + \frac{2\gamma_1}{\gamma_2 + \gamma_3 e^{2kQt}} \right)
\end{equation}
where $q_{\infty} = k/(k+1)$,  $k = C/ \sqrt{QR}$, $\gamma_1 = q_{\infty} - q_0$, $\gamma_2 = \gamma_1(k-1)$, $\gamma_3 = (k-1)q_0 - k$.

Notice that as $t \to \infty$, $q(t; q_0) \to q_\infty \leq 1$ monotonically. Thus $q_\infty$ defines the maximum attainable clarity. Equation~\eqref{eqn:clarity_dynamics_closed_form} can be inverted to determine the time required to increase the clarity from $q_0$ to some $q_1$. This time is denoted $\Delta T: [0, 1]^2 \to \Rnonneg$:
\eqn{
\Delta T(q_0, q_1) = t \text{ s.t. } q(t, q_0) = q_1 \quad \text{for } q_1 \in [q_0, q_\infty) \label{eqn:clarity_dynamics_inverse}
}
For $q_1 < q_0$, we set $\Delta T(q_0, q_1) = 0$ while $\Delta T(q_0, q_1)$ is undefined for $q_1 \geq q_\infty$. 


\vspace{-1mm}

%% file: subsections/problem_formulation.tex
\label{sec:problem_formulation}
\subsection{Environment Specification}
\begin{figure*}[t]
  \centering
  \includegraphics[width=2.0\columnwidth]{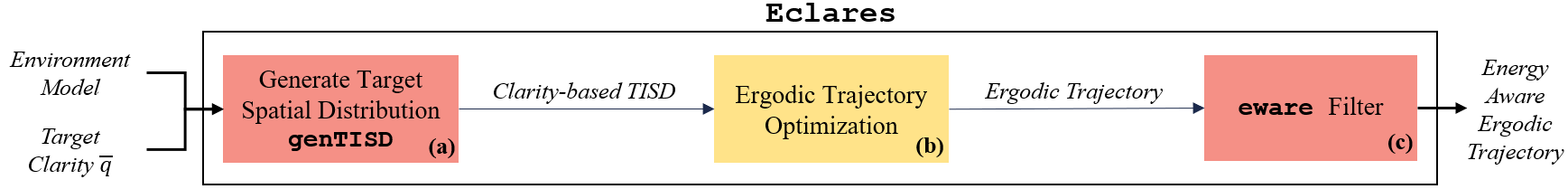}
  \caption{The high-level overview of the proposed framework. Blocks (a) and (c) are the proposed modules of the framework, while block (b) is borrowed from \cite{Dressel_Ergodic}.}
  \label{fig:eclares_overview}
\end{figure*}
Consider the coverage space $\Pcal$. We discretized the domain into a set of $N_c$ cells each with size $V$.\footnote{Size is width in 1D, area in 2D, and volume in 3D.} Let $m_c: [t_0, \infty) \to R$ be the (time-varying) quantity of interest at each cell $c \in \Ccal = \{1, ..., N_c\}$. We model the quantities of interest as independent stochastic processes:
\begin{subequations}
\label{eqn: quantity_of_interest}
\eqn{
    \Dot{m}_c &= w_{c}(t), &&w_{c}(t) \sim \mathcal{N} (0, Q_{c}) \label{eqn: quantity_of_interest_process}\\
    y_c &= C_{c}(x) m_{c} + v_c(t),  &&v_c(t) \sim \mathcal{N} (0, R) \label{eqn: quantity_of_interest_measurement}
}
\end{subequations}
where $y_c \in \R$ is the output corresponding to cell $c$. $R$ is the measurement noise, and $Q_c \in \Rplus$ is the process noise variance at each cell $c$. Since $m_c$ varies spatially and temporally under process noise $Q_c$ for each cell $c \in \Ccal$, the environment becomes a \emph{stochastic spatiotemporal environment}.

\subsection{Problem Statement}
\label{sec:motivating_example}
Consider a robotic system performing persistent coverage of a stochastic spatiotemporal environment~\eqref{eqn: quantity_of_interest}, i.e., over a time interval $[0, \infty)$. Assume the desired quality of information at each cell has been specified. This is encoded using a \emph{target clarity} $\overline q_c < q_{\infty, c}$ for each cell $c \in \Ccal$. The target clarity can be different at each cell, indicating a different desired quality of information at each cell, but must be less than $q_{\infty, c}$, the maximum attainable clarity of the cell. If $\overline q_c \geq q_{\infty, c}$ for a cell $c \in \Ccal$, then the robot would try to spend infinite amount of time at a cell $c$, which is undesirable.

We use clarity as our information metric since it is particularly effective for stochastic spatiotemporal environments:
\begin{itemize}
    \item The clarity decay rate in cell $c$, i.e. $-Q_{c}q_{c}^2$, is explicitly dependent on the stochasticity of the environment $Q_{c}$ in \eqref{eqn: quantity_of_interest}. This allows the information decay rate to be determined from the environment model, and not set heuristically~\cite{clarity}. Furthermore, spatiostatic environments are a special case: by setting $Q_{c} = 0$, clarity cannot decay. 
    \item While taking measurements of cell $c$, clarity $q_{c}$ monotonically approaches $q_{\infty, c} < 1$ for $Q_c, R > 0$. This indicates that maximum attainable information is upper bounded.
\end{itemize}

In this persistent task, the robot replans a trajectory every $T_H \in \Rplus$ seconds, i.e., at times $\{t_0, t_1, ... \}$ for $t_k = k T_H$, $k \in \naturals$. At the $k$-th iteration, the objective is to minimize the \textit{mean clarity deficit} $q_d(t_k + T_H)$, which is defined as
\begin{equation}
    \label{eq:mean_clarity_error}
    q_d(t_{k} + T_H) =  \frac{1}{N_c}\sum_{c = 1}^{N_c}  \max (0, \overline{q}_{c} - q_{c}(t_{k} + T_H)) 
\end{equation}
where $q_{c}(t_{k} + T_H)$ is the clarity at time $t_{k} + T_H$ of cell $c \in \Ccal$. However, in order to persistently monitor a stochastic spatiotemporal environment over a long time horizon, the robot's energy constraints must be taken into consideration. Thus, the overall problem can be posed as follows:
\begin{subequations}    
\label{eq:overall_prob}
\eqn{
\min_{\chi(t),u(t)} \quad &  q_d(t_{k} + T_H)\\
    \textrm{s.t.} \quad & \chi(t_k) = \chi_k\\
    & \Dot{\chi} = f(\chi, u)   \\
    & \Dot{q}_{c} = g(x, q_{c}) \ \forall c \in \Ccal  \\
    & e(t) \geq {e_{min}}   \label{eq:overall_prob_energy}
}
\end{subequations}
where $q_d(t_{k} + T_H)$ is the mean clarity deficit at the end of system trajectory $\chi(t; t_k, \chi_k)$, $\forall t \in [t_k, t_{k} + T_H]$ given by \eqref{eq:mean_clarity_error}, $g : \mathcal{X} \times [0, 1] \rightarrow \mathbb{R}_{\geq 0}$ define the clarity dynamics \eqref{eqn:clarity_dynamics}, and $e_{min}$ is the minimum energy level allowed for the robot.

%% file: subsections/Eclares.tex
\label{sec:ECLARES}

In this section, we introduce \Eclares{} as a solution to the optimization problem \eqref{eq:overall_prob}. We first provide the solution overview and discuss its motivation. Next, we provide details on the method to construct the TISD and lastly, we introduce the \eware{} filter. 

\subsection{Method Motivation \& Overview}

In order to solve problem~\eqref{eq:overall_prob}, we take inspiration from ergodic search~\cite{Mezic_Ergodic, Dressel_Ergodic}. As discussed in~\cref{sec:prelim_ergodic}, ergodic search can generate trajectories by solving problem~\eqref{eq:ergodic_optimzation_cont}. If the construction of TISD $\phi$ is based on the current clarity $q_c(t)$ and the target clarity $\overline{q}_{c}$ at each cell, then ergodic search can be used to minimize the mean clarity deficit~\eqref{eq:mean_clarity_error}. Therefore, in this work, we propose a method to construct $\phi$ using clarity. However, the ergodic trajectory optimization in \eqref{eq:ergodic_optimzation_cont} does not consider the energy constraint~\eqref{eq:overall_prob_energy}. One approach would be to add the energy constraint in \eqref{eq:ergodic_optimzation_cont}, but the non-convexity of \eqref{eq:ergodic_optimzation_cont} implies that guaranteeing convergence and feasibility is challenging. Therefore, we propose the framework \Eclares{}, shown in \cref{fig:eclares_overview}, as an approximate solution to the problem \eqref{eq:overall_prob}.\footnote{It is also important to note $q_d(T)$ is not differentiable so computing gradients for optimization problem \eqref{eq:overall_prob} can be challenging. However, the problem~\eqref{eq:ergodic_optimzation_cont} is differentiable and can be approximately solved using PTO~\cite{Dressel_Ergodic}.} 
\begin{algorithm}
\caption{The \genTISD{} algorithm}\label{alg:genTSD}
\begin{algorithmic}[1]

\Function{ \genTISD{} }{$q_{c}$, $\overline{q}_{c}$, environment model \eqref{eqn: quantity_of_interest}}
    \For{$c \in \{1, ..., N_c\}$}
        \State $k \gets 1/ \sqrt{Q_{c}R}$
        \State $q_{\infty} \gets k/(k+1)$
        \State $\overline{q} \gets \min (\overline{q}_{c} , q_{\infty, c} - \epsilon)$
        \State $\phi_c \gets \Delta T(\overline{q}, q_c)$
    \EndFor
    \State $\phi_{c} \gets \phi_{c}/({\sum_{c = 1}^{N_c} \phi_{c}}), \quad \forall c \in \{1, ..., N_c\}$
    \State \Return $\phi_{c} \  \forall c \in \{1, ..., N_c \}$
\EndFunction
\end{algorithmic}
\end{algorithm}
\begin{algorithm}
\caption{The $\eware$ algorithm}\label{alg:eware}
\begin{algorithmic}[1]

\Function{$\eware$}{$\tau_j, x_k^{ergo}, x_{j-1}^{com}$}
    \State Solve back-to-base trajectory \eqref{eq:b2b_prob}
    \State Solve initial value problem \eqref{eqn:candidate}
    \State $\tau_f \gets \tau_j + T_N + T_B$
    \If{ $e(t) \geq 0 \ \forall t \in [\tau_j, \tau_f]$  and $x(\tau_f) = x_c$}
        \State $x_j^{com} \gets x_j^{can}$
    \Else
        \State $x_j^{com} \gets x_{j-1}^{com}$
    \EndIf
    \State \Return $x_j^{com}$
\EndFunction
\end{algorithmic}
\end{algorithm}
Our solution involves decoupling~\eqref{eq:overall_prob} into two sub-problems: (A) we design a trajectory that maximizes information collection without the energy constraint (referred to as the \emph{ergodic trajectory}), and (B) we construct a trajectory that tracks part of the ergodic trajectory but also reaches the charging station before the robot's energy depletes (referred to as the \emph{committed trajectory}). The low level controller of the robot always tracks the last committed trajectory. In this way, we ensure that the robot persistently explores the domain without violating the energy constraint.  

We run these steps at different rates. The ergodic trajectory is replanned every $T_H$~seconds, while the committed trajectory is updated every $T_E < T_H$ seconds.\footnote{$T_E, T_H \in \Rplus$ are user-defined parameters.} That is,
\begin{itemize}
    \item at each time $t_k = k T_H, k \in \naturals$, 
    \begin{itemize}
        \item recompute the TISD $\phi$ using \genTISD{}. 
        \item recompute an ergodic trajectory using PTO \cite{Dressel_Ergodic}. 
    \end{itemize}
    \item at each time $\tau_j = j T_E, j \in \naturals$, 
    \begin{itemize}
        \item update the committed trajectory using $\eware$.
    \end{itemize}
\end{itemize}

In this work, we adopt PTO directly from \cite{Dressel_Ergodic}. 

Before describing \genTISD{} and $\eware$, we establish notation for trajectories. Let $x([t_k, t_k + T_H]; t_k, x_k)$ be the ergodic trajectory generated at time $t_k$ starting at state $x_k$ and defined over a time horizon $T_H$. For compactness, we refer to $x([t_k, t_k + T_H]; t_k, x_k)$ as $x^{ergo}_{k}$. We use similar notation to describe other trajectories later on.

\subsection{Generate Target Spatial Distribution (\genTISD{})}

The \genTISD{} algorithm is described in \cref{alg:genTSD}. Let $\phi_c$ denote the target information density evaluated for cell $c$. At the $k$-th iteration (i.e, at time $t_k = k T_H$), we set $\phi_c$ to be the time that the robot would need to increase the clarity from $q_c(t_k)$ to the target $\overline q_c$ by observing cell $c$ (Lines 3-6). This is determined using~\eqref{eqn:clarity_dynamics_inverse}. The small positive constant $\epsilon > 0$ in Line 5 ensures that target clarity is always less than the maximum attainable clarity, i.e., $\overline q_c < q_{\infty, c}$. Finally, we normalize $\phi_c$ such that the sum of $\sum_{c \in \Ccal} \phi_c = 1$ (Line 8). Once $\phi$ is constructed, PTO method \cite{Dressel_Ergodic} is used to generate the ergodic trajectory.

\begin{figure} [t]
  \centering
  \includegraphics[width=1.0\columnwidth]{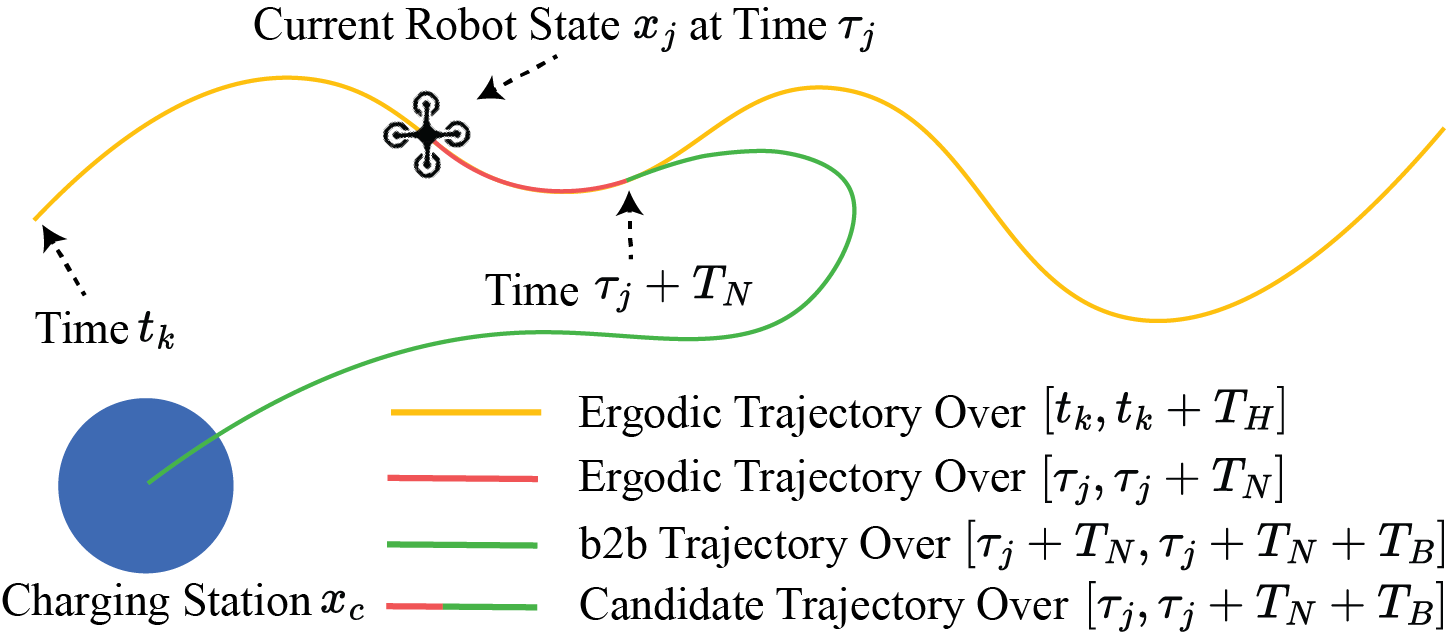}
  \caption{This figure illustrates the generation of candidate trajectory at time $\tau_j$, combining a small portion of the ergodic trajectory (shown in red) and the b2b trajectory (shown in green).}
  \label{fig:traj_eware}
\end{figure}
\begin{figure*}[t]
    \centering
    \includegraphics[width=1.0\linewidth]{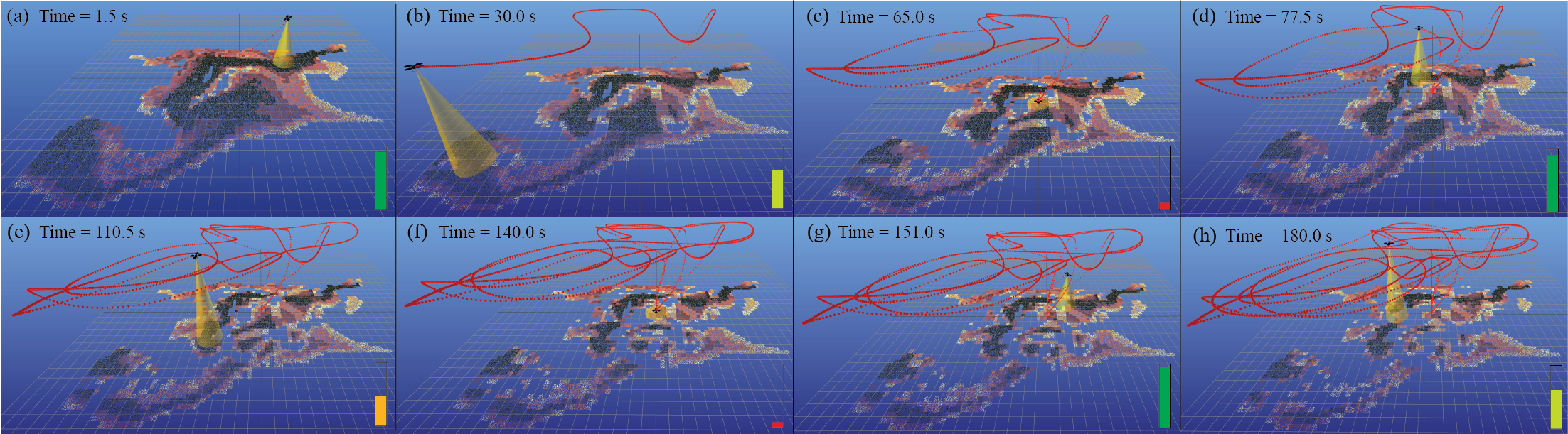}
    \caption{The still frames from quadrotor simulation are shown where the quadrotor with a downward-facing camera and a gimbal is persistently monitoring a stochastic spatiotemporal environment. The TISD is shown as a point cloud in each frame and the colored bar in each frame shows the battery level. Frame (a) shows the quadrotor starting the coverage mission with full energy. Moving across frames (b) and (c), the quadrotor energy depletes and returns to the charging station (modeled as the origin of the world frame). After battery swapping, the quadrotor resumes the coverage mission as shown in the frame (d). The same trend follows in the subsequent frames. Moreover, it can be seen that as the quadrotor explores the environment, the TISD becomes more sparse, which reflects increase in clarity in the region. For full simulation video, please see \textcolor{purple}{https://youtu.be/i2yD3A2b5nA}}
    \label{fig:Quad_Still_Frames_2}
\end{figure*}

\subsection{Energy-Aware ($\eware$)}
\label{sec:eware}
Inspired by~\cite{gatekeeper}, $\eware$ is a lightweight filter to ensure that the robot reaches the charging station before running out of energy. To do so, it constructs a \textit{candidate} trajectory that follows the ergodic trajectory for a short time horizon $T_N$ before attempting to return to the charging station. If the candidate trajectory is valid (defined below), it becomes a \textit{committed} trajectory. The low-level tracking controller always tracks the last committed trajectory. \cref{alg:eware} describes the $\eware$ algorithm.

The $j$-th iteration of \eware{} starts at time $\tau_j = j T_E$. Let the robot state at time $\tau_j$ be $x_j \in \Xcal$ and the system system at time $\tau_j$ be $\chi_j \in \Zcal$. Suppose the last ergodic trajectory generated is $x_{k}^{ergo}$ (defined over the interval $[t_k, t_k + T_H]$). 

(Line 2) We construct a back-to-base (b2b) trajectory $x_{j}^{b2b}$, defined over $[\tau_j + T_N, \tau_j + T_N + T_B]$, by solving the optimal control problem:
\begin{subequations}
\label{eq:b2b_prob_overall}
\eqn{
        \min_{x(t),u(t)} \ &  \int_{\tau_j + T_N}^{\tau_j + T_N + T_B} \norm{x(\tau) - x_{c}}_{\mathbf{Q}}^2 +  \norm{u(\tau)}_{\mathbf{R}}^2 d\tau \label{eq:b2b_prob}\\
        \textrm{s.t. } \ & x(\tau_j + T_N) = x_{j}^{ergo}(\tau_j + T_N), \\
        &  \Dot{x} = f_r(x, u) 
}
\end{subequations}
where $x_c \in \Xcal$ is the state of the charging station, $\mathbf{Q} \in \pd^{n}$ weights state cost, and $\mathbf{R} \in \pd^m$ weights control cost.

(Line 3): Once b2b trajectory $x_{j}^{b2b}$ is generated, we numerically construct the candidate trajectory by solving the initial value problem 
\begin{subequations}
\eqn{
\label{eqn:candidate}
    \Dot{\chi} &= f(\chi, u(t)),\\
    \chi(\tau_j) &= \chi_j \\
    u(t) &= \begin{cases} \pi(\chi,  x_{k}^{ergo}(t)), & t \in [\tau_j, \tau_j + T_N) \\
        \pi (\chi,  x_{j}^{b2b}(t)), &  t \in [\tau_j + T_N, \tau_j + T_N + T_B]
   \end{cases} 
}
\end{subequations}
where $\pi : \mathcal{Z} \times \mathcal{X} \rightarrow \mathcal{U}$ is a control policy to track the portion of the ergodic trajectory and the b2b trajectory. \cref{fig:traj_eware} illustrates the generation of a candidate trajectory.

(Lines 5-9): Once the candidate trajectory is constructed, we check whether the robot can reach the charging station before running out of energy. If it can, the candidate replaces the committed trajectory.

%% file: subsections/experiments.tex
\section{Simulation Case Study}
\label{sec:experiments}
\subsection{Simulation Setup}
In this section, we evaluate \textbf{\texttt{Eclares}} using a simulation case study. We synthesize a hypothetical scenario where a quadrotor modeled with nonlinear dynamics described in \cite[Eq. (10)]{jackson2021planning} is persistently monitoring a stochastic spatiotemporal environment modeled according to the dynamics described in \eqref{eqn: quantity_of_interest}. The coverage domain is 20$\times$20 meter in area with grid cells of dimension 0.20$\times$0.20. We use the PTO method \cite{Dressel_Ergodic} with double integrator dynamics to solve the optimization problem \eqref{eq:ergodic_optimzation_cont} for the time horizon $T_H = 30.0$s. To generate b2b trajectory, we solve the problem \eqref{eq:b2b_prob_overall} using model predictive control (MPC) with reduced linear quadrotor dynamics proposed by \cite{jackson2021planning}. Another MPC problem with the same reduced linear quadrotor dynamics was posed to track the ergodic trajectory for shorter time horizon $T_N = 2.0$s. Then the committed trajectory is generated using the \eware{} algorithm. The committed trajectories are generated in the receding-horizon manner at  $0.50$ Hz (i.e., $T_E = 2.0$s) and tracked at $20.0$ Hz with zero-order hold.\footnote{For more details on algorithm implementation, trajectory generation, and on hypothetical stochastic environment, please see https://github.com/kalebbennaveed/Eclares.git.} If one of the committed trajectory returns quadrotor back to the charging station, it stays there to simulate recharging. Note that no assumptions are made about the charging method; it may either involve stationary charging or battery swapping. We use the battery dynamics described in \cite[Eq. (6)]{persis_cbf3} as it directly incorporates robot control input as opposed to the worst-case approximation used in other works \cite{persis_cbf1, persis_cbf4}. Throughout the simulation, we use the Runge-Kutta 4th order (RK4) integration scheme. \cref{fig:Quad_Still_Frames_2} shows still frames from the light UAV simulator where a quadrotor with downward-facing camera is exploring a stochastic spatiotemporal environment.  
\begin{figure*}[t]
    \centering
    \includegraphics[width=1.0\linewidth]{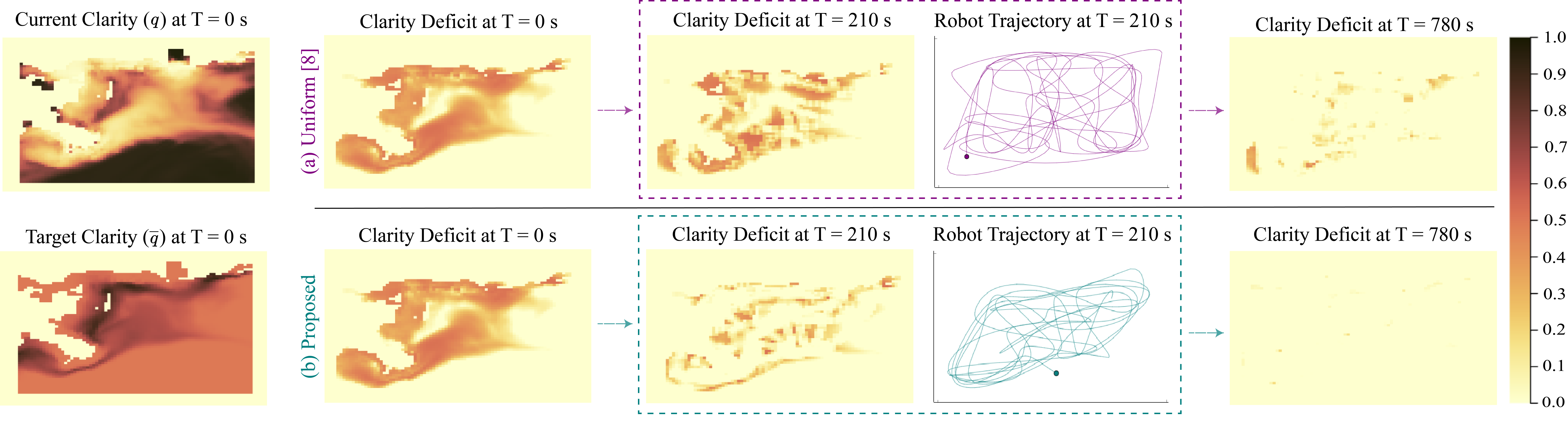}
    \caption{The clarity deficit Map for the \textbf{spatiostatic environment} at time T = 0 s, T = 210 s, and T = 780 s is shown comparing two methods, uniform distribution, and proposed method, with the same clarity $q$ and target clarity $\overline{q}$ at T = 0 s.}
    \label{fig:clarity_deficit_map}
\end{figure*}
\begin{figure*}[t]
    \centering
    \includegraphics[width=1.0\linewidth]{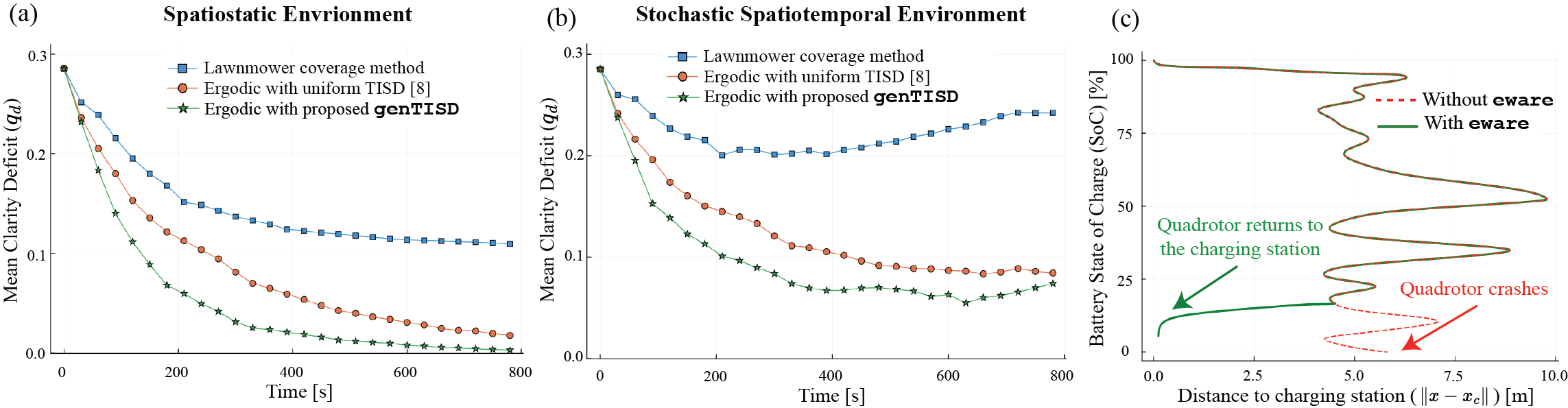}
    \caption{(a) Plot of the mean clarity deficit $(q_d)$ against time is shown for the \textbf{spatiostatic environment} for three methods.  (b) The plot of the $q_d$ against time is shown for the \textbf{stochastic spatiotemporal environment} for three methods. It can be seen from both plots that the proposed method outperforms other methods: Ergodic search with uniform target spatial distribution and classic lawnmower coverage method. Moreover, it is important to note that in the stochastic spatiotemporal environment, it is not possible to acquire zero clarity error as information is always decaying due to term $-Qq^2$ in \eqref{eqn:clarity_dynamics}. (c) Plot of the quadrotor's distance to charging station $(\|x - x_c\|)$ against battery's State of Charge (SoC) is shown for one quadrotor flight, where $x_c$ is the position of the charging station. In case of \eware{}, quadrotor safely returns to the charging station as energy depletes with almost  3 \% SoC. However, without \eware{}, quadrotor unaware of its remaining energy keep on following the nominal path, leading to a crash.
    }
    \label{fig:clarity_deficit_domain}
\end{figure*}
\subsection{Results and Discussion}
\subsubsection{\textbf{Performance Comparison between Spatiostatic and Stochastic Spatiotemporal Environment}}
\cref{fig:clarity_deficit_domain}(a) and \cref{fig:clarity_deficit_domain}(b) show the evolution of mean clarity deficit $q_d$ for the spatiostatic and stochastic spatiotemporal environments respectively. For the proposed method, in the case of the spatiostatic environment, the $q_d$ approached almost zero at time T = 780s; however, for the stochastic spatiotemporal environment, the $q_d$ reduced to a non-zero value. This is expected due to the inherent stochasticity present in the environment as the information is decaying constantly due to the term $-Qq^2$ in \eqref{eqn:clarity_dynamics}, and it is not possible to achieve zero clarity deficit. However, the proposed method does provide an ability to set realistic expectations on the maximum possible information that can be collected in a stochastic environment by computing $q_{\infty, c}$, $\forall c \in C$.
\subsubsection{\textbf{Comparison to Baseline methods}}
We compare the performance of the proposed methodology \genTISD{} to the uniform TISD method used in most of the ergodic literature \cite{Mezic_Ergodic, dong2023time}. The uniform TISD assigns equal importance to all cells $c \in \Ccal$ in the domain. As evident from  \cref{fig:clarity_deficit_domain}(a) and \cref{fig:clarity_deficit_domain}(b), the proposed method shows a faster convergence rate as compared to the uniform method. This is explained by the robot trajectory shown in \cref{fig:clarity_deficit_map} as the robot spends more time in the region with higher clarity deficit as compared to the uniform method. It is important to note that the past ergodic literature does not consider the stochastic nature of the environment in their construction of target distribution. We also evaluate the performance of the proposed coverage method against the classic lawnmower coverage method. In the lawnmower coverage method, the robot follows the prescribed path which covers the whole domain. The lawnmower coverage pattern results in a repetitive paths as the robot covers same areas multiple times while leaving gaps in some areas. This causes clarity deficit to stabilize to a non-zero value as shown in \cref{fig:clarity_deficit_domain}(a). In case of using lawnmower coverage method for stochastic spatiotemporal environment, the clarity deficit eventually starts increasing as the information constantly decays due to the term $-Qq^2$ in \eqref{eqn:clarity_dynamics}. 

\subsubsection{\textbf{Evaluation of \texttt{eware} Filter}}
\cref{fig:clarity_deficit_domain}(c) compares the quadrotor flight with and without the \eware{} filter. Evidently, with the \eware{} filter, the quadrotor safely returns to the charging station. However, in the absence of \eware{} filter, quadrotor unaware of its remaining battery energy keeps on following the nominal ergodic trajectory, leading to a crash. Moreover, we observed that although the minimum SoC specified was $0\%$, the quadrotor returns to the charging station with almost $3 \pm 1$ \% remaining SoC. This number can be brought down by increasing the frequency of the candidate trajectory generation from $0.5$ Hz to a larger number; however, this comes at the expense of overall computational time. We also 
compared the compute time of proposed method \eware{} with that of the CBF-based methods used for robotic task persistence since both are online methods. The \eware{} filter, updated at 0.5 Hz, took average time of 30.2 ms to numerically forward integrate the candidate trajectory through quadrotor and battery's nonlinear dynamics. Alternatively, the CBF-based method used in \cite{persis_cbf1}, updated at 20 Hz, took average time of 0.0551 ms to solve the optimization problem described in \cite[Eq. (2)]{persis_cbf1} with single integrator dynamics. While considering the update frequency, the CBF-based method \cite{persis_cbf1} is almost 13 times faster than the proposed method; however, it assumes simple robot and battery models. Conversely, the proposed \eware{} filter achieved real-time performance with nonlinear quadrotor and battery dynamics. This allows for more flexibility with robot and battery models and thus increases generalizability of the proposed method . 
\vspace{-3.0mm}


%% file: subsections/conclusion.tex
\section{Conclusion}%
In this work, we introduced a framework \textbf{\texttt{Eclares}}, in which we proposed a method to update target information spatial distribution for the ergodic trajectory optimization, which considers the stochastic spatiotemporal environments. We also integrated a lightweight filter called \textbf{\texttt{eware}} between the high-level ergodic planner and low-level tracking controller, which iteratively ensures that the robot has enough battery energy to continue the coverage task. Through simulation case study, we demonstrated that the proposed framework incorporates potential information decay and quantifies maximum attainable information in a stochastic environment while ensuring robot's safe return to the charging station before the battery energy depletes. However, We note that the decoupling used in \textbf{\texttt{Eclars}} can lead to a suboptimal solution, since the b2b trajectory is not optimized for information collection, which we hope to address in future work. Other future directions include considering environmental disturbances and extension of the proposed method to multi-agent exploration scenarios.

%% file: subsections/Appendix.tex
\label{sec:appendix}

\subsection{System Dynamics Model}

\subsubsection{3D Quadrotor Dynamics}
We modeled the quadrotor using the dynamics described in \cite[equation (10)]{jackson2021planning}. The dynamics have the form:


\begin{equation}
\label{eq:quad_dynamics}
    \Dot{x} =
    \begin{bmatrix}
        \Dot{r} \\
        \Dot{q} \\
        \Dot{v} \\
        \Dot{\omega}
    \end{bmatrix}
    =
     \begin{bmatrix}
        v \\
        (\frac{1}{2})L(q)H\omega  \\
        (\frac{1}{m}) ^WF(x,u) \\
        J^{-1}(^B\tau(x,u) - \hat{\omega}J\omega)
    \end{bmatrix}   
\end{equation}

where $x \in \mathbb{R}^{13}$ and $u \in \mathbb{R}^{4}$ are the state and control vectors, $r \in \mathbb{R}^3$ is the position vector, $q \in \mathbb{H}$ is attitude where $\mathbb{H}$ is space of quaternions, $v \in \mathbb{R}^3$ is the linear velocity, $\omega \in \mathbb{R}^3$ is the angular velocity, $m$ is the mass, $J \in \mathbb{R}^{3 \times 3}$ is the inertia matrix ,$^WF(x,u) \in \mathbb{R}^3$ are the forces in the world frame, $^B\tau(x,u)$ are the moments in the body frame, $H \in  \mathbb{R}^{4\times3}$ is the matrix used to convert the 3-dim vector into 4-dim quaternion with zero scalar part, 

\begin{equation}\
    H = 
    \begin{bmatrix}
        0.0 & 0.0 & 0.0 \\
        1.0 &  0.0 & 0.0 \\
        0.0 &  1.0 & 0.0 \\
        0.0 &  0.0 & 1.0
    \end{bmatrix}
\end{equation}

$L(q) \in \mathbb{R}^{4\times4}$ is the matrix used for left multiplication of quaternion, 
\begin{equation}\
    L(q) = 
    \begin{bmatrix}
        q_s & -q_v^T \\
        q_v &  q_s I + \hat{q_v} \\
    \end{bmatrix}
\end{equation}

where $q_s \in \mathbb{R}$ is the scalar part of the quaternion, $q_v \in \mathbb{R}^3$ is the vector part of the quaternion, and $\hat{x}$ is the skew-symmetric matrix operator

 \begin{equation}
    \hat{x} = 
    \begin{bmatrix}
        0 & -x_3 & x_2\\
        x_3 & 0 & -x_1\\ 
        -x_2 &  x_1 & 0\\  
    \end{bmatrix} 
\end{equation}


\subsubsection{Battery Dynamics}

We modeled the battery dynamics using the dynamics described in \cite[equation (6)]{persis_cbf3}. The dynamics have the form: 

\begin{equation}
    \Dot{e} = -\eta C^{-1}\alpha(\| u\|^2)
\end{equation}

where $C$ is battery's rated capacity, $\eta$ is its Columbic efficiency, $\alpha$ is a class $\mathcal{K}$ function, and $\| u\|$ is the magnitude of the control input. 

\subsection{Ergodic Trajectory Generation}

We use the Projection-based Trajectory Optimization (PTO) proposed by \cite{Dressel_Ergodic} for ergodic trajectory optimization. For implementation sake, discrete approximation of system dynamics~\eqref{eqn:aug_ctrlaffine_sys} is used. The optimization problem to generate the $T_H$ horizon discrete trajectory $\tilde{x}_{1:N_{h}}$ is defined as

\begin{equation}
    \label{eq:ergodic_optimzation_discont}
    \begin{aligned}
        \min_{\tilde{x}_{1:N_{h}},\tilde{u}_{1:N_{h} - 1}} \quad & \{\Phi(\tilde{x}_{1:N_{h}}, \phi) + J_b(\tilde{x}_{1:N_{h}}) + \frac{\Delta t}{2}\sum_{k = 1}^{N_{h}-1} \tilde{u}_k^T\mathbf{R}\tilde{u}_k \}\\
    \textrm{s.t.} \quad & \tilde{x}_{1} = \tilde{x}_{ic}, \\
    & \tilde{x}_{k+1} = \tilde{x}_{k} + f(\tilde{x}_{k}, \tilde{u}_k)\Delta t
    \end{aligned}
\end{equation}

where  $\mathbf{R} \in \pd^m$ weights control cost, $\tilde{x}_{ic} $ is the initial robot state, and $J_b(\tilde{x})$ is the boundary penalty to keep the robot within the boundary of the rectangular coverage space $\mathcal{P}$. The $J_b(\tilde{x})$ is given as 

\begin{equation}
    J_b(\tilde{x}_{1:N_{h}}) = c_b \sum^{N_h}_{k = 1} \sum^s_{i = 1} (\max (\tilde{x}_{k,i} - L_i, 0)^2 + \min (\tilde{x}_{k,i}, 0)^2)
\end{equation}

where $c_b > 0$ weights the boundary penalty, $L_i$ is the $i^{th}$ length of the rectangular domain, and $\tilde{x}_{k,i}$ is the $i^{th}$ element of state $\tilde{x}_{k}$. We used the double integrator dynamics while solving the problem \eqref{eq:ergodic_optimzation_discont}. Comprehensive details on the PTO method can be found in \cite{Dressel_Ergodic}. We iteratively generate ergodic trajectories of time horizon $T_H = 30$ sec with $dt = 0.2$ at a frequency of $\frac{1}{30}$ Hz. If the quadrotor returns to the charging station while following the nominal trajectories due to energy depletion, the new ergodic trajectory is generated based on the current clarity level after recharging. 
\subsection{Candidate Trajectory Generation}

\begin{figure*}[t]
    \centering
    \includegraphics[width=1.0\linewidth]{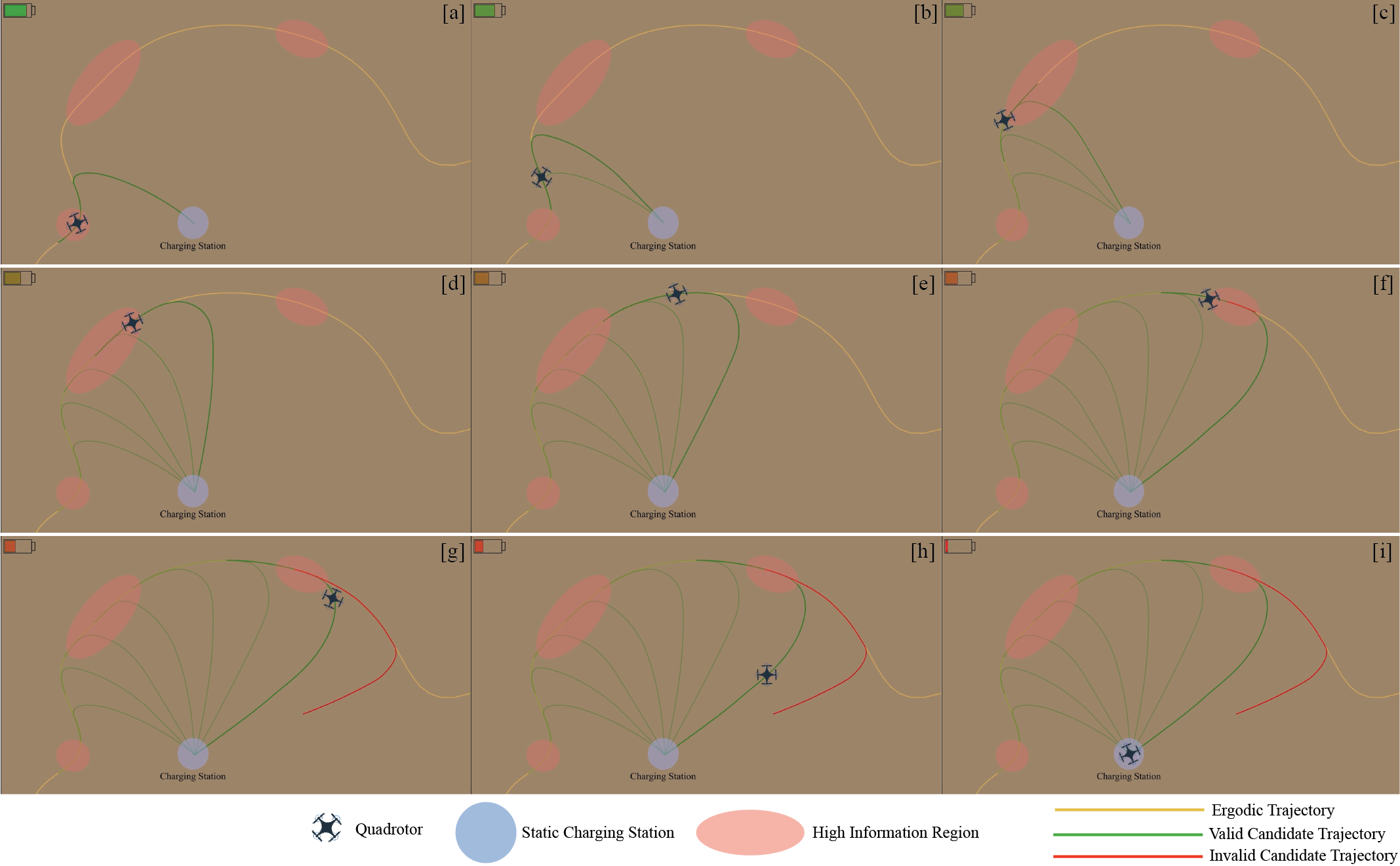}
    \caption{This figure illustrates the process of generating candidate trajectories iteratively. The green trajectories are the valid candidate trajectories or the committed trajectories. These trajectories follow part of the ergodic trajectory (shown in yellow) and then go to the charging station. If the candidate trajectory cannot ensure the quadrotor's safe return to the charging station, it is considered invalid (shown in red) and the quadrotor takes the last committed trajectory back to the charging station.}
    \label{fig:eware_seq}
\end{figure*}

The candidate trajectories are constructed by using the method described in \cref{sec:eware}. The candidate trajectories are generated at $0.5$ Hz and tracked with zero-order hold at $20.0$ Hz. Essentially, the candidate trajectories are generated by stitching the shorter $T_N$ horizon ergodic trajectory with the $T_B$ horizon b2b trajectory. Here, we describe the method to generate the $T_N$ horizon ergodic trajectory where $T_N < T_H$ and the $T_B$ horizon b2b trajectory. 

\subsubsection{\textbf{$T_N$ horizon Ergodic Trajectory}}

The shorter horizon ergodic state trajectory $\tilde{x}_{1:N_n}$ and the control trajectory $\tilde{u}_{1:N_{n}-1}$ are generated using the convex Model Predictive Controller (MPC) formulation given as

\begin{equation}
    \label{eq:nominal_traj}
    \begin{aligned}
        \min_{\tilde{x}_{1:N_n},\tilde{u}_{1:N_{n}-1}} \quad &  \sum_{k = 1}^{N_n-1} \Bigl[ (\tilde{x}_k - \tilde{x}^{ref}_{k})^T\mathbf{Q}(\tilde{x}_k - \tilde{x}_{k}^{ref}) \\ + 
 & (\tilde{u}_k - \tilde{u}_{k}^{ref})^T\mathbf{R}(\tilde{u}_k - \tilde{u}_{k}^{ref}) \Bigr]\\
    \textrm{s.t.} \quad & \tilde{x}_{1} = \tilde{x}_{ic}, \\
    & \tilde{x}_{k+1} = f(\Bar{x}_{k}, \Bar{u}_{k}) + \Biggl[\frac{\partial f}{\partial x}\bigg|_{\Bar{x}_{k}, \Bar{u}_{k}}\Biggr](\tilde{x}_k - \Bar{x}_{k}) \\
    & \hspace{1.0cm} + \Biggl[\frac{\partial f}{\partial u}\bigg|_{\Bar{x}_{k}, \Bar{u}_{k}}\Biggr](\tilde{u}_k - \Bar{u}_{k})
    \end{aligned}
\end{equation}

where $(\Bar{x}_{k}, \Bar{u}_{k})$ is the $k^{th}$ equilibrium point, $ \mathbf{Q} \in S_{++}^{m}$ weights state cost, $ \mathbf{R} \in S_{++}^{m}$ weights control effort, $\tilde{x}_{k}^{ref}$  is the $k^{th}$  reference state from the reference $T_H$ horizon ergodic state trajectory $\tilde{x}^{ref}_{1:N_n}$, $\tilde{u}_{k}^{ref}$ is the $k^{th}$ reference control input from $T_H$ horizon ergodic control trajectory $\tilde{u}^{ref}_{1:N_n}$, and $N_n = \frac{T_N}{dt}$. For the convex MPC problem, we use the linearized version of the nonlinear quadrotor dynamics described in \eqref{eq:quad_dynamics}. It is important to note that the linearized quadrotor system with $A \in \mathbb{R}^{13 \times 13}$ and $B \in \mathbb{R}^{13 \times 4}$ is not controllable with rank 12. Therefore, we reduce the system to $\Bar{A} \in \mathbb{R}^{12 \times 12}$ and $\Bar{B} \in \mathbb{R}^{12 \times 4}$ using the attitude Jacobin for error dynamics proposed by \cite{jackson2021planning}. For more details, please see \cite{jackson2021planning}. The reference state trajectory and reference control trajectory used in the  MPC problem \eqref{eq:nominal_traj} are formed using the double integrator ergodic trajectory with unit quaternion. Apart from this method, differential flatness property can be used to track the double integrator trajectory using the nonlinear geometric controller \cite{lee2010geometric}. We use the values of $T_N = 2.0$ sec with sampling time $dt = 0.05$.
\subsubsection{\textbf{back-to-base(b2b) Trajectory}}

The b2b robot state trajectory $\tilde{x}_{1:N_b}$ and control trajectory $,\tilde{u}_{1:N_{b}-1}$ are generated using the convex MPC formulation given as

\begin{equation}
    \label{eq:b2b_traj}
    \begin{aligned}
        \min_{\tilde{x}_{1:N_b},\tilde{u}_{1:N_{b}-1}} \quad &  \sum_{k = 1}^{N_b-1} \Bigl[ (\tilde{x}_k - x_{c})^T\mathbf{Q}(\tilde{x}_k - x_{c}) + 
 (\tilde{u}_k)^T\mathbf{R}(\tilde{u}_k ) \Bigr]\\
    \textrm{s.t.} \quad & \tilde{x}_{1} = \tilde{x}_{N_h}, \\
    & \tilde{x}_{k+1} = f(\Bar{x}_{k}, \Bar{u}_{k}) + \Biggl[\frac{\partial f}{\partial x}\bigg|_{\Bar{x}_{k}, \Bar{u}_{k}}\Biggr](\tilde{x}_k - \Bar{x}_{k}) \\
    & \hspace{1.0cm} + \Biggl[\frac{\partial f}{\partial u}\bigg|_{\Bar{x}_{k}, \Bar{u}_{k}}\Biggr](\tilde{u}_k - \Bar{u}_{k})
    \end{aligned}
\end{equation}

where $x_c$ is the state of the charging station and the initial state $\tilde{x}_{1}$ is the last state of the $T_N$ horizon ergodic trajectory. We use the same reduced linearized quadrotor dynamics to generate a b2b trajectory. We use the values of $T_B = 10.0$ sec with sampling time $dt = 0.05$.

\cref{fig:eware_seq} illustrates the process of generating candidate trajectories iteratively. The green trajectories are the valid candidate trajectories or the committed trajectories. These trajectories follow part of the ergodic trajectory (shown in yellow) and then go to the charging station. If the candidate trajectory cannot ensure the quadrotor's safe return to the charging station, it is considered invalid (shown in red) and then the quadrotor takes the last committed trajectory back to the charging station. 

\subsection{Construction of the hypothetical stochastic spatiotemporally varying environment}

We constructed a hypothetical stochastic spatiotemporally varying environment using the Salinity data from the Gulf of Mexico \cite{Gulf_data}. Salinity, one of the oceanographic quantities, refers to the concentration of dissolved salts in water. In the oceanographic community, salinity evolution in a water body can be used to study cross-shelf exchanges. Cross-shelf exchanges refer to the movement of nutrients from the shallow waters to the open ocean. We extracted the salinity mean values along with the standard deviation for a specific region and then scaled down the environment spatially and temporally to form a hypothetical environment. 




